# GraphCodeBERT for Cross-Language Code Clone Retrieval through Data-Flow Alignment and Hard-Negative Mining

Qiong Liu [1*]

[1] Moutai Institute,, Renhuai, 564507, Guizhou, China

**Abstract**

Cross language code clone retrieval supports software migration, maintenance, and the discovery of implementations that share behavior across programming languages. Lexical variation obscures equivalent computations, while superficially similar programs create difficult distractors in embedding retrieval. This paper proposes a GraphCodeBERT adaptation that combines soft data flow alignment with reliability filtered hard negative mining. The alignment component compares variable dependency representations across paired implementations without requiring identical identifiers or graph topology. The mining component excludes known equivalent solutions, retains a mixture of random and challenging candidates, and gradually increases negative difficulty. A reproducible evaluation design separates programming tasks across training, validation, and testing and covers six directed retrieval settings involving Java, Python, and Go. Across the six settings, the joint configuration has mean reciprocal rank 0.814, compared with 0.731 for a matched GraphCodeBERT baseline. Removing alignment reduces this score to 0.786, whereas replacing filtered mining with random negatives reduces it to 0.772. The evaluation framework also addresses retrieval depth, perturbation tolerance, mining contamination, and indexing cost under a fixed candidate protocol. The component comparison separates structural supervision from negative selection and identifies their respective contributions to the ranking profile. The resulting design provides an auditable route for investigating whether language independent dependencies and carefully selected distractors improve cross language code retrieval under realistic annotation and computational constraints. The absolute mean reciprocal rank increase is 0.083, corresponding to approximately 11.4% relative to the baseline; this aggregate comparison does not establish statistical significance.

*Keywords:* Cross Language Clone Retrieval, Data Flow Alignment, Hard Negative Mining, GraphCodeBERT Adaptation, Contrastive Representation Learning

## 1. Introduction

Cross-language clone detection research combines enhanced abstract syntax trees with graph neural networks to represent structural and semantic differences between Java and Python functions [1]. A systematic review classifies code similarity methods by their machine learning algorithms, code representations, datasets, and evaluation metrics [2]. Scalable clone search combines BERT-based code embeddings with nearest-neighbor retrieval to reduce the cost of searching for syntactically divergent clones [3]. In the related task of source code summarization, ESALE uses summary-focused multi-task learning to strengthen alignment between code representations and natural-language summaries [4]. These studies address complementary representation and retrieval concerns, while cross-language function search additionally requires an explicit account of behavioral relevance. Software repositories frequently contain implementations of related functionality in different programming languages. A service may be migrated from Java to Go, an analysis routine may be prototyped in Python before being deployed elsewhere, and an algorithm may be independently implemented by several development teams. Finding corresponding functions can reduce the effort needed to inspect migration changes, investigate duplicated maintenance responsibilities, and identify alternative implementations. The same computation can use different library calls, iteration conventions, exception mechanisms, and container representations. A text index can therefore miss a useful implementation even when its control decisions and value transformations closely match the query. Conversely, two functions may share descriptive identifiers and common operations while differing in boundary handling or output semantics. Function retrieval is particularly sensitive to ranking quality because a developer usually inspects only a short-ranked list. A relevant implementation placed deep in that list contributes little practical value. The present problem concerns retrieving candidate functions written in a different programming language, with relevance defined by documented task equivalence and supported, where possible, by executable evidence. This definition deliberately avoids treating identical problem labels as a complete proof of equivalence. Input assumptions, numerical behavior, and failure modes can remain different even when two implementations are associated with the same task.

Pretrained source code encoders offer a useful starting point because they represent tokens in their computational context. Cross language transfer and preservation of retrieval specific relations remain distinct representational requirements. Language dependent syntax may dominate the distance between two functions, especially when the training corpus contains uneven language coverage or repeated repository conventions. GraphCodeBERT incorporates data flow into contextual source code representations. Recent structure-aware adaptation methods combine code graph representations with language models through graph-text alignment, as illustrated by CGP-Tuning for vulnerability detection [5]. Dependency-aware representations offer a bridge between implementations whose surface forms differ.

Structural correspondence remains challenging because equivalent functions can distribute computations across different temporary variables, combine several operations into a single expression, or invoke a library routine in place of an explicit loop. A rigid node correspondence would penalize these transformations. Tree-based semantic clone detection extracts structural patterns beyond literal token identity, as demonstrated by the Goner representation [6]. A retrieval model therefore needs a mechanism that encourages compatible dependency roles without assuming that every node has a unique counterpart. It also needs an explicit policy for incomplete graphs. Confidence in structural supervision should reflect parse failure, unresolved calls, and function truncation through an explicit availability policy.

The examples used to train a retrieval encoder matter as much as the representation supplied to it. Random negative candidates often differ so strongly from a query that they contribute limited discrimination after early training. In binary code similarity research, diversity-sensitive contrastive learning compares candidate functions across compilation conditions to improve representation discrimination [7]. Hard candidates can expose confusions involving shared identifiers, similar loops, or related operations, and the validity of each high scoring negative requires separate verification. Incomplete annotations may label an alternative correct implementation as irrelevant. Evidence from federated clone detection further identifies skewed data distributions and restricted access to industrial code as barriers to transferring benchmark performance into practice [8].This paper investigates a joint design in which soft data flow alignment supplies structural supervision and reliability filtered mining supplies discriminative pressure. The contribution is an adaptation of an existing encoder, combining local dependency supervision and reliability-controlled candidate selection within one retrieval pipeline rather than introducing a new pretraining architecture. The research questions concern their separate effects, their interaction, and the operational cost of maintaining a candidate pool. Section 2 positions the problem relative to representation learning, structural matching, and negative sampling. Section 3 defines the retrieval setting and the proposed alignment mechanism. Section 4 specifies mining, optimization, and serving procedures. Section 5 develops the experimental protocol and analyzes retrieval performance, including ablations and deployment tradeoffs. Section 6 identifies the contribution of the design, its limitations, and the work required for empirical validation.

## 2. Related Work

### 2.1. Cross language representation learning

Cross language clone retrieval differs from monolingual duplicate detection because the relationship between lexical overlap and shared implementation history varies across languages. Methods based on token normalization reduce naming variation but retain language specific operators, library conventions, and statement boundaries. Learned encoders seek a common representation space by exposing a model to multiple programming languages or by training it on aligned examples. Du et al. [9] introduced AdaCCD, which uses adaptively refined contrastive learning to transfer clone detection representations from resource-rich programming languages to languages without annotated clone data. The key modeling choice is the unit of correspondence: a whole function, a statement fragment, or a task level family. Function supervision is economical, but it can hide local mismatches that become consequential during maintenance. Task supervision provides many positives, yet programs solving the same task may embody different assumptions or output conventions. These distinctions affect the interpretation of retrieval success.

Cross-consistency training leverages cross-lingual similarity to train code language models for clone detection and code search [10]. Such consistency can improve transfer when direct paired labels are limited, while behavioral correctness requires additional verification. An encoder can be consistently sensitive to unhelpful features if those features recur across the supervision pool. Evaluation therefore needs task disjoint partitions, language specific reporting, and comparison under a common candidate set. The proposed design retains a shared encoder to control representational capacity, while adding explicit local structural supervision and a documented mining policy. Its intended contribution is a testable combination of local structural supervision and explicit negative selection within a shared representation space. Whether the combination generalizes beyond curated algorithmic functions remains an empirical question.

### 2.2. Structural supervision and dependency matching

Structural encoders use relations among program elements to complement token context. Abstract syntax captures grammatical construction, whereas data flow approximates dependencies through which values are created and consumed. Program dependency graph approaches can connect graph nodes through code keywords to improve semantic clone detection and identify corresponding lines [11]. Such dependency information is relevant to retrieval because two implementations can preserve value propagation despite differences in identifier spelling. However, extracted dependency graphs are partial static descriptions. Aliasing, external libraries, dynamic dispatch, and language dependent parser support affect what relations are available. A structural match provides evidence of computational similarity whose behavioral implications require semantic verification.

Graph-based code search methods learn global and fine-grained alignments to represent similarity between queries and source code [12]. Local alignment is relevant when one language introduces temporary variables and another uses nested expressions. A weighted correspondence can absorb these differences, although unconstrained matching may collapse onto common operations and overlook rare but decisive computations. The present design therefore combines learned node similarity with role compatibility, edge context, and graph availability masks. It supports flexible correspondence cardinality and many to many node relations. This choice separates cross language correspondence from identical graph reconstruction. The proposed evaluation must still test whether gains arise from meaningful dependencies, extra parameters, or incidental graph size signals. Capacity matched controls, shuffled edges, and explicit reporting of extraction coverage are needed to distinguish these explanations. Such controls expose parser quality and support assessment of preprocessing related selection bias.

### 2.3. Negative sampling and annotation uncertainty

Negative sampling determines which nonmatching candidates receive training pressure. Uniform sampling is simple and reproducible, but many sampled functions become uninformative as representation quality improves. Hard negative code search methods select semantically challenging candidates and include their representations in gradient-based contrastive optimization [13]. This adaptability creates a feedback loop: early model errors influence which examples appear difficult, and subsequent updates can reinforce those errors. A sampler that always chooses the nearest candidate can overrepresent ambiguous tasks or mislabeled equivalents. Diversity and reliability therefore need separate treatment from raw similarity.

Liu et al. [14] proposed an uncertainty-aware code search approach that combines diverse query and code embeddings with hard negative sampling to improve contrastive representation learning. Its methodological relevance concerns sampling reliability across natural language code search and code to code retrieval. Cross language clones introduce different sources of ambiguity, including shared task families and language specific test coverage. The proposed filtering policy uses only evidence available in the training partition and preserves a random component to avoid concentrating every update on a narrow neighborhood. Rank based selection also limits dependence on the calibration of similarity scores during early training. These code-search findings motivate the sampling design, but their query modality differs from paired source-code retrieval. Their reported effectiveness is not evidence that the present model outperforms them under a common benchmark.

Inter-dataset code duplication can contaminate evaluation when fine-tuning test examples overlap with pretraining data [15]. These considerations motivate an evaluation that records the negative pool, exposes the rejection reasons, and treats unresolved candidate pairs separately from verified negatives. The remaining gap is an auditable investigation of local dependency alignment and reliability filtered mining within a common GraphCodeBERT retrieval setting. A factorial comparison separates component contributions, interaction effects, and uncertainty in their joint behavior. The cited studies cover recent clone detection and adjacent code-search settings. They establish methodological context; differences in query modality, supervision, and candidate construction prevent their published scores from serving as direct baselines here.

## 3. Data-Flow-Aligned Retrieval Representation

### 3.1. Retrieval units and graph construction

The input consists of a query function and a collection of candidate functions written in another programming language. Each function is associated with its source text, parser status, and task family identifier where such metadata is available. Task identifiers support supervised partitioning and training filters, while inference relies on source representations and parser metadata. Relevance is defined over function pairs through the evaluation annotation protocol. Multiple candidates may be relevant to a query, and their source repositories may differ. The retrieval system returns a ranked list for candidate inspection. A separate threshold can support clone triage, with precision and recall calibrated on validation data drawn from the intended candidate distribution. Task-family relevance is an annotation rule for retrieval, whereas behavioral equivalence additionally requires compatible input domains, outputs, and failure behavior. A high rank under the family rule therefore supports candidate discovery rather than a proof of clone equivalence.

Preprocessing preserves operators, literals, and library names while removing comments and nonfunctional formatting under the primary protocol. A language specific parser extracts variable definitions, uses, and approximate def-use edges. Each graph node links to the subword span representing its corresponding variable occurrence. The preprocessing record includes unresolved references, graph size, truncation status, and parse diagnostics. Functions exceeding the token budget retain an entry window and dependency related spans according to a deterministic selection rule. All structural configurations follow the same rule. Functions with failed parses remain in the evaluation set and receive token-based encoding together with an explicit structural availability mask. This treatment preserves the benchmark population across preprocessing conditions. Reproducing this preprocessing policy requires the parser release, grammar revision, extraction rules, and subword-to-node mapping. These implementation identifiers are not specified here, so the policy description alone is insufficient to reproduce graph extraction exactly. The retained-span policy must additionally specify token ordering, entry-window allocation, and tie resolution among dependency spans; the current budgets alone do not determine the selected input.

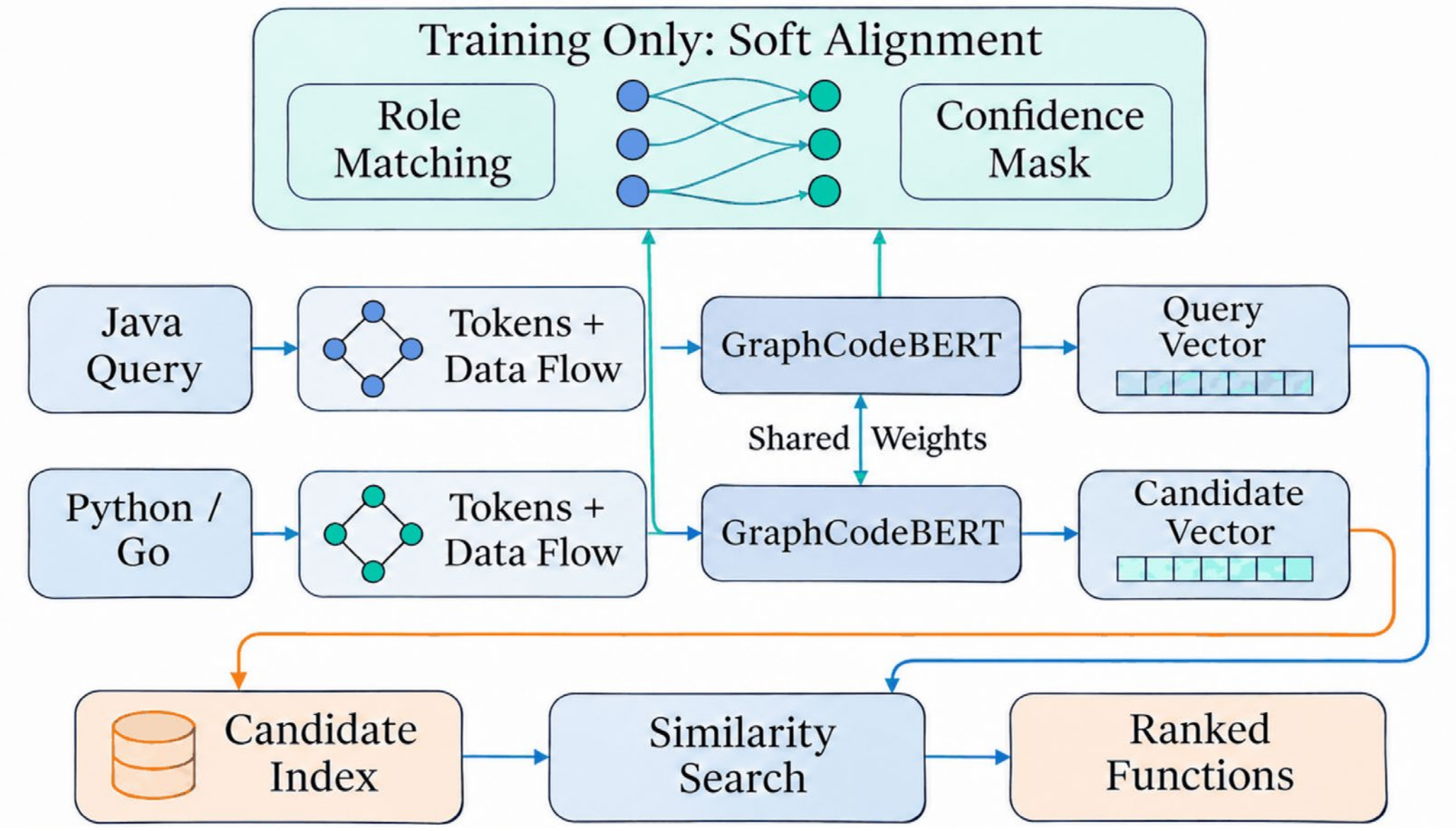

**Figure 1. Proposed retrieval architecture with language specific parsing, shared GraphCodeBERT encoding, soft dependency alignment during training, pooled embeddings, and indexed cross language retrieval**

The architecture in Figure 1 separates structural supervision during training from candidate retrieval during serving. Source and target functions pass through shared encoder weights, while their parser adapters remain language specific. Token states and variable node states are pooled separately before projection into a normalized retrieval space. Alignment gradients update the shared representation during training. At inference, each candidate is encoded independently and stored in an index. Pairwise graph correspondence therefore contributes to representation learning, while indexed vector comparison determines the ranked retrieval output. Runtime evaluation accounts separately for structural training overhead and query processing cost.

### 3.2. Soft correspondence across dependency roles

Each available graph node receives a contextual encoder state and a compact role description derived from its incoming and outgoing dependencies. Roles distinguish input associated values, intermediate transformations, condition related values, and output associated values where static evidence permits. A coarse role vocabulary accommodates differences among language front ends. Role compatibility contributes a bounded bias to the correspondence distribution. Learned projections place source and target node states in a common space, allowing their affinities to reflect contextual similarity alongside dependency roles. A null matching option accommodates source variables whose target counterpart is absent or structurally unresolved. The following equation converts these affinities into directional correspondence weights:

$$a_{ij} = \frac{exp(c_{ij}/\tau_a)}{\sum_{k \subset V_t \cup \{\emptyset\}} exp(c_{ik}/\tau_a)} \tag{1}$$

Here, $a_{ij}$ denotes the weight linking source node i to target node j, $c_{ij}$ combines projected similarity with a bounded role bias, and $\tau_a$ controls correspondence concentration. The target set $V_t$ includes the available target nodes, while $\emptyset$ denotes the learned null state. Correspondence weights are calculated for paired training functions with usable graphs. A reverse alignment pass balances supervision across language dependent node granularities. Node states are normalized before affinity calculation to control the influence of representation magnitude. Graph availability masks determine which examples contribute structural supervision, while retrieval supervision remains active for all eligible pairs.

The alignment penalty compares each source state with its correspondence weighted target representation. A weak dependency consistency term additionally encourages compatible neighborhoods under soft matching. Its contribution remains limited because accumulator expansion, temporary elimination, and expression restructuring can change local graph topology while preserving computation. Confidence weights decrease when extraction is incomplete or correspondence mass concentrates on the null state. The alignment equation incorporates this confidence weighted residual together with dependency consistency:

$$\mathcal{L}_{align} = \frac{\sum_i q_i \left\| u_i - \sum_j a_{ij} v_j \right\|_2^2}{\sum_i q_i} + \beta \mathcal{L}_{edge} \tag{2}$$

In this equation, $u_i$ and $v_j$ are projected source and target node states, respectively, and $q_i$ weights the reliability of each source node. The coefficient β controls neighborhood consistency. The reverse direction contributes an analogous loss, and the two directional values are averaged. A pair with zero valid confidence contributes zero structural loss. Gradients pass through the projections and contextual encoder states, while parser decisions remain fixed. Separate controls for the null bias, role bias, and edge term help identify their respective effects. Inspection of selected correspondences supports failure diagnosis alongside retrieval evaluation over the complete test partition. Null correspondence represents unavailable structural support rather than an irrelevant function pair. When total node confidence is zero, the implementation must bypass the residual division and return zero structural loss while retaining retrieval supervision. The neighborhood term requires an explicit edge comparator and coefficient before implementation-level reproduction. Its separate effect is not isolated by the reported ablations, which compare alignment as a complete component.

### 3.3. Retrieval pooling and scoring

The function representation combines a contextual sequence summary with an availability weighted graph summary. Attention pooling operates over valid variable nodes, while masking excludes padding positions. A learned gate conditions on graph availability and normalized node count. The sequence branch remains active for every function. When parsing fails, the graph contribution becomes zero and the projection receives a dedicated availability indicator. This provides a consistent fallback path whose frequency can be examined during evaluation. A capacity matched control retains the projection and gate while disabling structural supervision, allowing alignment effects to be distinguished from changes in parameter capacity. The availability indicator records whether structural input exists; it is not a relevance label. Zero graph contribution leaves the sequence representation available to the shared projection, so a failed parse changes the representation path rather than removing the function from retrieval.

Function pooling and candidate scoring are expressed by a shared retrieval equation:

$$\begin{aligned} z_f &= normalize\left(W_p\left[h_f; g_f p_f; m_f\right]\right), \\ s(q,c) &= z_q^{\top} z_c. \end{aligned} \tag{3}$$

The sequence summary $h_f$, pooled graph state $p_f$, gate $g_f$, and availability indicator $m_f$ determine the normalized function vector $z_f$. The projection matrix $W_p$ maps the concatenated representation into the retrieval space. The scoring component of the equation calculates the dot product between normalized query and candidate vectors. Mining, validation, and serving use the same pooling procedure so that candidate difficulty remains consistent across stages. Language metadata selects parser adapters, while a common projection preserves a shared similarity geometry across retrieval directions.

Exact dot product search supports the primary effectiveness comparison. Approximate indexing is evaluated separately as candidate volume increases, with explicit reporting of index recall and latency. Storage accounting includes vectors, identifiers, index overhead, and retained preprocessing artifacts. Reducing vector dimensionality changes both representational capacity and serving cost, so each reduced dimension configuration requires matched comparison conditions. Independent candidate encoding also permits index updates when repository contents change. Encoder versioning remains essential because a consistent representation space underlies every similarity calculation within an index.

## 4. Reliability Filtered Mining and Optimization

### 4.1. Candidate selection and curriculum

Mining begins with embeddings computed exclusively from the training partition. For each query, the sampler retrieves a neighborhood larger than the final negative set. Known positive task family members, exact duplicates, and pairs with verified equivalence evidence are excluded. Unresolved near duplicates enter a quarantine set for evidence review. Training identifiers, embeddings, and relevance annotations determine candidate selection. Each quarantine record preserves the reason for withholding a candidate, supporting later inspection of annotation gaps. Reinstatement requires additional evidence from the training partition. This boundary maintains the independence of training candidate selection and test evaluation.

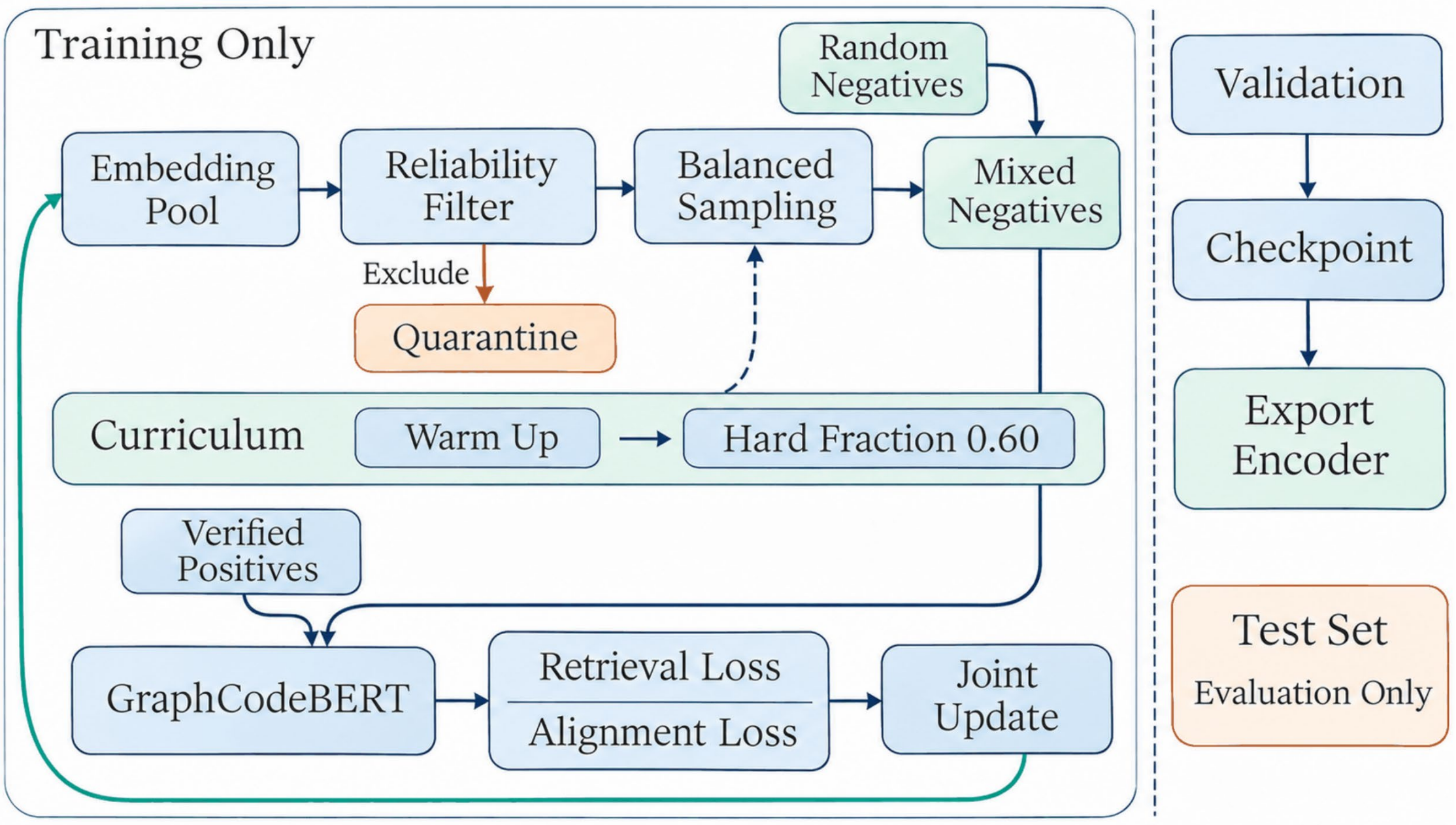

**Figure 2. Training workflow for pool refresh, training only exclusion filters, uncertainty quarantine, rank band sampling, mixed negative batches, joint optimization, and validation-based stopping**

Figure 2 shows how candidate selection enters the optimization cycle. A warm-up period uses random candidates to establish an initial retrieval geometry. Subsequent epochs combine random negatives with candidates sampled across a configurable rank band. The hard negative fraction increases gradually until it reaches its configured ceiling. Retaining a random component preserves coverage of the broader candidate collection, while language balanced sampling controls the contribution of each target language. The primary protocol refreshes the embedding pool once per epoch. A separate systems comparison varies the refresh interval while keeping optimizer updates constant. Logging candidate ranks, acceptance decisions, and language composition allows changes in retrieval performance to be examined alongside changes in the training distribution. The hard fraction refers to accepted negatives entering the batch, not the raw retrieved neighborhood. If filtering leaves too few eligible hard candidates, random eligible negatives fill the remaining slots; quarantined pairs remain excluded.

### 4.2. Joint retrieval and structural objectives

For each query, verified positive counterparts contribute to a multi-positive contrastive objective. The denominator contains these positives and the accepted negative set. Rejected and quarantined candidates remain outside the optimization set. Reliability weights scale the contribution of accepted negatives according to the strength of their supporting evidence. Batch composition is recorded because mining changes the effective negative distribution throughout training. A fixed temperature controls score concentration during the primary comparison, and normalized function vectors maintain consistency between optimization and retrieval. These requirements lead to the retrieval loss equation:

$$\mathcal{L}_{ret} = -log\left[\frac{\sum_{p\subset P_q} exp(s(q,p)/\tau)}{\sum_{p\subset P_q} exp(s(q,p)/\tau) + \sum_{n\subset N_q} w_n exp(s(q,n)/\tau)}\right] \tag{4}$$

The positive set $P_q$ contains verified counterparts available to query q, while $N_q$ contains accepted negatives. The temperature $\tau$ determines the concentration of similarity scores. Each accepted negative receives a reliability weight $w_n$ bounded away from zero. Query level averaging gives each retrieval instance equal weight despite differences in positive multiplicity. Comparing uniform and graded negative weights helps distinguish changes in reliability treatment from changes in effective update magnitude. Symmetric training reverses eligible language pairs and averages their directional losses. Numerically stable accumulation is retained for normalization and logarithmic operations when mixed precision is used. Acceptance determines membership in the negative set, whereas reliability weighting changes the contribution of an accepted candidate. Table 2 does not report a separate uniform-versus-graded weight comparison or specify numerical reliability calibration. Because the numerator aggregates positive mass, one strongly scored positive can reduce the query loss even when other relevant candidates rank poorly. Multi-relevant recall is therefore needed alongside first-relevant-rank metrics to assess coverage of the positive set.

Structural supervision is introduced after warm-up, when projected node states provide a more stable basis for correspondence. Its coefficient increases over a short schedule and then remains fixed. The alignment schedule and maximum hard negative fraction are controlled separately so that their effects remain interpretable. Early stopping follows validation mean reciprocal rank averaged across the predefined language directions. Parser coverage, mining acceptance, and directional ranking statistics provide supplementary diagnostics. The complete optimization objective combines retrieval discrimination, structural alignment, and regularization through the equation below:

$$\mathcal{L} = \mathcal{L}_{ret} + \lambda\mathcal{L}_{align} + \gamma||\theta_{new}||_2^2 \tag{5}$$

Within the objective equation, λ controls structural supervision and γ regulates the newly introduced projection and gating parameters, denoted by $\theta_{new}$. The shared encoder receives gradients from both retrieval and alignment losses. Optimizer weight decay applied to pretrained parameters is recorded separately from this additional regularizer. Gradient clipping limits large updates arising from difficult batches, and each optimizer update requires at least one valid supervised query. Checkpoint selection is performed independently for each seed under the same update budget. Validation results determine the alignment coefficient, negative curriculum, and stopping point.

### 4.3. Training records and serving export

Reproducibility depends on preserving the complete training configuration and its associated records. These records include partition identifiers, preprocessing versions, initialization hashes, random seeds, candidate pools, rejection reasons, and validation decisions. Epoch level diagnostics report graph availability, accepted hard candidates, quarantine counts, and average positive multiplicity. These quantities help distinguish representation changes from changes in which queries receive supervision. A capacity matched baseline retains the proposed projection and pooling while disabling alignment and replacing the mining distribution. A token-based control additionally examines the contribution of extracted structure. Together, these controls support attribution of performance differences to specific modeling choices. Table 2 reports the baseline, joint model, alignment-only, filtered-mining-only, and unfiltered-mining configurations. Separate capacity-matched token controls and shuffled-edge results are not supplied, so attribution to dependency semantics remains conditional on those additional comparisons.

The exported encoder contains the modules required to produce function vectors. Correspondence tensors, training queues, and quarantine metadata remain in the training records. Candidates are preprocessed once per collection revision and embedded using a versioned checkpoint. Each index contains vectors generated by the same encoder version. Query parsing failures activate the token-based fallback used during training. Serving logs record fallback and truncation status to support drift analysis. Triage thresholds are versioned with the encoder and calibrated against the intended deployment distribution. Changes to preprocessing, pooling, or projection require corresponding index regeneration to preserve a consistent retrieval space. An incremental candidate update is valid only when encoder, tokenizer, parser, pooling, and projection versions remain fixed. A change to any representation-defining component requires re-encoding the candidate collection before mixing old and new entries.

Training cost includes parsing, forward and backward computation, correspondence construction, and embedding pool refresh. Serving cost comprises query parsing, encoding, index access, and optional verification of returned candidates. These stages are accounted for separately so that lookup performance and end to end response time remain distinguishable. Structural correspondence contributes to training overhead, while independent vector encoding determines the serving path. Runtime comparisons specify the device, batch size, concurrency level, software stack, candidate collection, and timing procedure. Recording these conditions supports comparisons across deployment configurations and identifies whether resource demand arises primarily from preprocessing, representation encoding, or index access.

## 5. Experimental Evaluation and Analysis

### 5.1. Evaluation protocol and controlled conditions

This section specifies the evaluation conditions and examines retrieval effectiveness, component behavior, robustness, and resource requirements. The corpus specification contains Java, Python, and Go functions organized into task families, with six directed cross language retrieval settings. The CoSQA+ benchmark uses test-driven agents to verify multiple suitable code implementations for a query, addressing limitations of annotation based solely on code inspection [16]. The present task-family protocol specifies 800 training families, 100 validation families, and 100 test families, each with ten functions per language. Thus, the partition specification contains 24000, 3000, and 3000 functions, respectively. Families are disjoint across partitions. Repository and author grouping should be enforced where provenance permits, and known translations or generated variants must remain with their source family. Table 1 records the intended conditions. The partition sizes provide comparable language coverage while preserving separation among task families. These counts specify the evaluation population but do not identify its source corpus. A reproducible release additionally requires source locations, collection dates, licenses, family mappings, and the provenance checks used to enforce repository grouping.

**Table 1. Evaluation design and controlled conditions**

| Item | Configuration | Control |
|---|---|---|
| Languages | Java, Python, Go | Six directions |
| Families | 800 / 100 / 100 | Task disjoint |
| Functions | 24000 / 3000 / 3000 | 10 per language per family |
| Test candidates | 1000 per direction | Complete target pool |
| Relevant candidates | 10 per query | Family rule |
| Token / graph budget | 384 / 128 | Matched across models |
| Seeds | 11, 12, 13, 14, 15 | Same update budget |
| Epoch / batch limit | 10 / 32 queries | Validation stopping |

The CoIR benchmark evaluates code information retrieval models across diverse retrieval tasks and programming languages [17]. For the present cross-language evaluation, each direction uses 1000 query functions and 1000 target functions from the test partition, with ten relevant candidates per query under the task family rule. Every query is evaluated against the complete target collection, and each relevant candidate retains its rank. Mean reciprocal rank uses the first relevant position; recall at a cutoff uses the proportion of all relevant candidates retrieved. With ten relevant candidates, recall at one has an upper bound of one tenth. The top one statistic is therefore success at one, the fraction of queries whose first candidate is relevant. Recall at ten and recall at fifty retain their conventional multi relevant meaning. These definitions maintain consistency between the relevance protocol and the reported ranking metrics. Candidate order ties are resolved by stable identifiers. Per query records include the number of relevant candidates, the first relevant rank, and relevant candidate counts at each cutoff. These records permit separate checks of ranking quality and annotation coverage. Queries with an empty relevance set require an explicit exclusion reason and a separately reported count. For task level aggregation, query scores are averaged within each family before family level summaries are calculated. This ordering makes the weighting policy transparent when future collections contain unequal numbers of implementations per family. For the balanced design, query scores are averaged within a direction and the six directional means receive equal weight in the overall score. Family summaries describe heterogeneity separately and should not be substituted for this directional macro-average.

Research on type-specific masking evaluates pretrained code models across several downstream tasks and examines the influence of pretraining objectives on their performance [18]. The baseline and joint configuration share the same GraphCodeBERT initialization, token budget of 384, graph budget of 128 nodes, vector dimension of 256, batch size of 32 queries, and maximum of ten epochs. The optimizer configuration uses learning rate 0.00002 for the encoder and 0.0001 for added modules. Five seeds, numbered 11 through 15, define the repeat protocol. The alignment coefficient is selected from 0.05, 0.10, 0.15, 0.20, and 0.30 using validation data only. Negative filtering and graph alignment are varied independently in the factorial comparison. The selected configuration uses an alignment coefficient of 0.15 and a maximum hard fraction of 0.60. Parameter selection follows the validation criterion specified in the optimization protocol.

Task families form the proposed bootstrap unit, preserving within family dependence among retrieval observations. Actual execution should resample the same families for both configurations and retain paired score differences. Seed variation and family sampling uncertainty describe different sources of variability and should be reported separately. Recent cross-lingual clone evaluation compares generative language models with embedding-based classifiers on common datasets, exposing substantial differences among model families [19]. The analysis distinguishes aggregate score profiles from estimates of sampling uncertainty. A complete reporting record comprises per query rankings, seed level outputs, graph coverage logs, and hardware timing traces. Paired differences preserve the relationship between configurations on the same family, while stratification by language direction identifies uneven variation. Confidence intervals and significance assessments require the underlying repeated observations and a prespecified resampling procedure. The aggregate values reported below are not accompanied by seed-level dispersions or paired confidence intervals. They support descriptive comparisons; statistical significance and consistency across seeds remain unestablished without the underlying repeated rankings.

### 5.2. Retrieval behavior and component effects

Table 2 presents the central retrieval comparison. Its six directional scores give a baseline mean reciprocal rank of 0.731 and a joint score of 0.814, an absolute difference of 0.083. The strongest joint setting is Java to Python at 0.842, whereas Go to Python is 0.791. Cross-domain code search research identifies limited labeled data and domain-specific semantics as obstacles to transferring pretrained retrieval models [20]. Interpreting directional variation requires consideration of extraction coverage, representation quality, and the composition of relevant candidates. Direction specific reporting exposes a spread of 0.051 among joint settings and identifies where aggregate performance conceals variation. The table also separates random negative replacement, alignment removal, and unfiltered hard mining, which achieve aggregate scores of 0.772, 0.786, and 0.794. The comparison keeps the candidate collection, preprocessing budget, and ranking criterion fixed across configurations. The six absolute directional gains range from 0.078 to 0.091. Their common sign shows that the reported aggregate improvement is not driven by a single direction, while the unequal gains leave language-specific extraction and annotation effects open to investigation.

**Table 2. Directional retrieval scores and ablation means**

| Condition | J→P | P→J | J→G | G→J | P→G | G→P | Mean |
|---|---|---|---|---|---|---|---|
| Baseline | 0.764 | 0.752 | 0.735 | 0.724 | 0.711 | 0.700 | 0.731 |
| Joint | 0.842 | 0.831 | 0.819 | 0.808 | 0.793 | 0.791 | 0.814 |
| Alignment only | 0.802 | 0.793 | 0.778 | 0.766 | 0.750 | 0.743 | 0.772 |
| Filtered mining only | 0.821 | 0.808 | 0.791 | 0.778 | 0.765 | 0.753 | 0.786 |
| Unfiltered mining | 0.823 | 0.809 | 0.800 | 0.788 | 0.774 | 0.770 | 0.794 |

The retrieval profiles connect ranking quality to the candidate inspection budget: Figure 3(a) shows directional mean reciprocal ranks from 0.700 to 0.842; Figure 3(b) gives recall at cutoffs 10, 20, 30, 50, and 100, with baseline values 0.682, 0.761, 0.804, 0.853, and 0.901 and joint values 0.758, 0.826, 0.865, 0.908, and 0.946; Figure 3(c) reports task family mean reciprocal rank quartiles whose joint median is 0.86 against 0.73 for the baseline; Figure 3(d) shows success at one decreasing from 0.704 to 0.420 as the joint model faces larger candidate collections. The collection size profile uses independently sampled distractor additions, and its values are ordered from the largest to the smallest collection. CCDive combines abstraction with local sequence alignment to recognize modified code and identify transformation locations, motivating evaluation under varied similarity conditions [21]. The profiles address complementary aspects of retrieval: directional rank concerns transfer, recall concerns coverage, the box plot concerns heterogeneity, and collection size concerns competition. Success at one and mean reciprocal rank capture different regions of the ranking distribution. The former measures immediate relevance, while the latter also credits relevant candidates appearing at subsequent positions. Examining both identifies improvements in the first returned candidate alongside changes deeper in the ranked list.

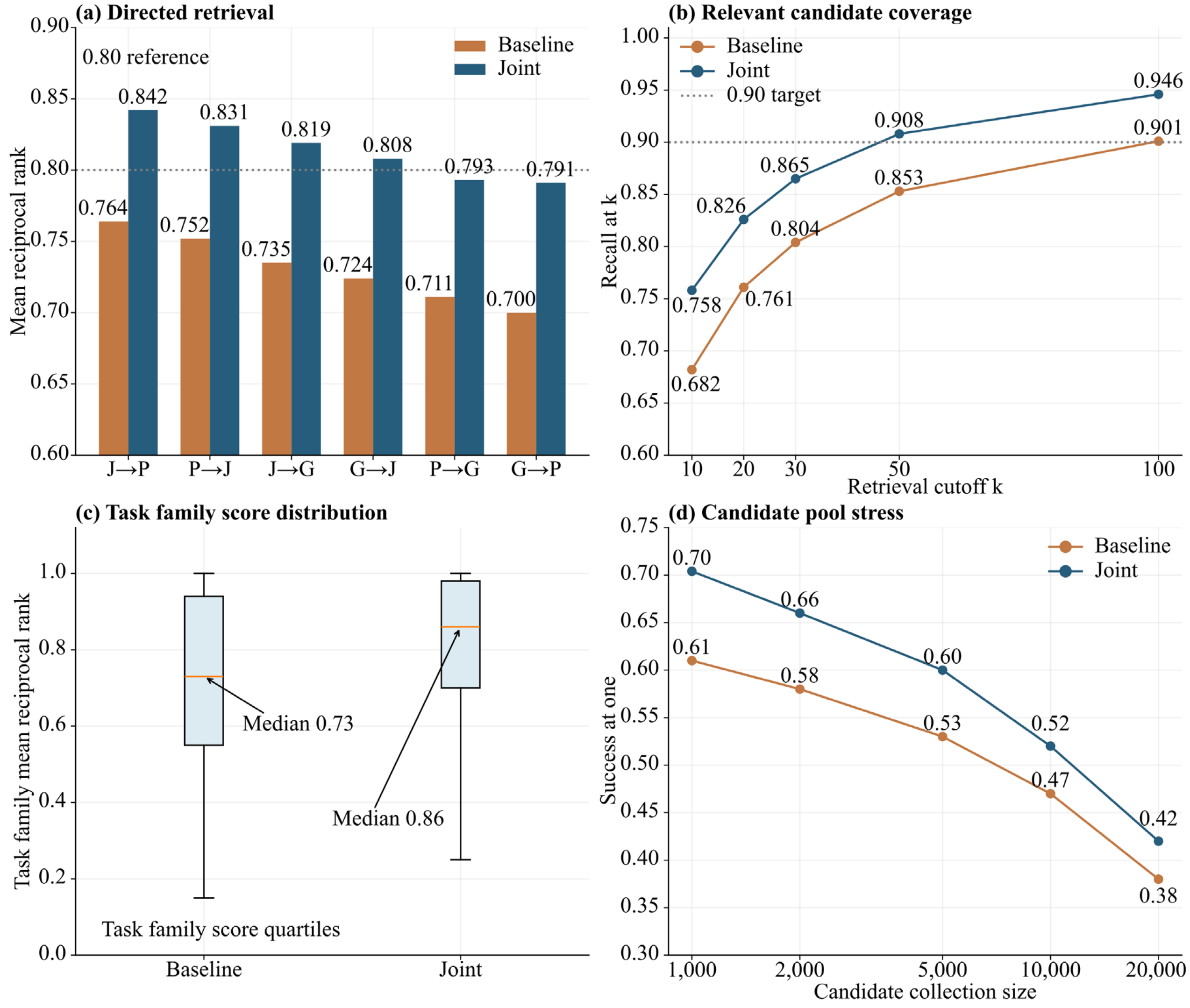

**Figure 3. Retrieval profiles: (a) paired directional score bars; (b) recall depth curves; (c) task family mean reciprocal rank box plots; (d) collection size success curves**

The component comparison examines interaction across the controlled configurations: Figure 4(a) places four factorial configurations at 0.731, 0.772, 0.786, and 0.814 and adds an unfiltered mining control at 0.794; Figure 4(b) shows the corresponding success at one value 0.610, 0.655, 0.673, 0.704, and 0.686. Naming-agnostic contrastive code search combines graph and path views of abstract syntax trees to reduce dependence on variable naming conventions [22]. Alignment alone contributes 0.041 and filtered mining alone contributes 0.055 relative to the baseline in the first metric. Their sum is 0.096, larger than the joint difference of 0.083, so the interaction is negative 0.013 on this scale. The negative interaction is consistent with overlapping contributions from alignment and mining on the mean reciprocal rank scale. The unfiltered control is 0.020 below the joint score, which defines the contrast intended to isolate reliability filtering. Additional shuffled edge and capacity matched token controls would distinguish sensitivity to meaningful dependency structure from sensitivity to added representational capacity. Their inclusion would refine attribution of the observed component contrasts. The joint configuration ranks above either individual component, but the negative interaction precludes describing the observed gain as superadditive synergy. This arithmetic contrast is descriptive and does not establish an interaction effect across repeated runs.

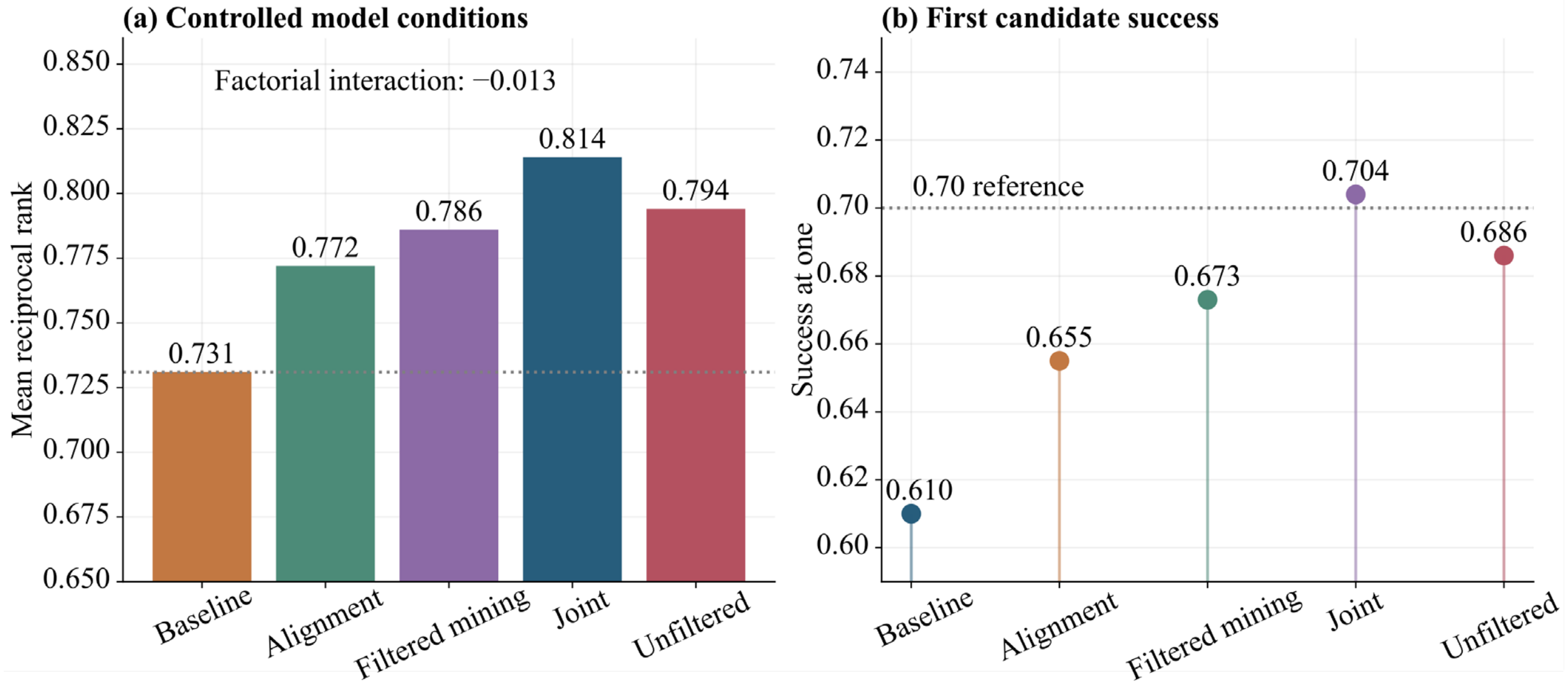

**Figure 4. Component effects: (a) score bars for five controlled configurations; (b) success at one dot profile for the same configurations**

### 5.3. Robustness sensitivity and operational cost

Robustness evaluation of Transformer-based code models uses semantics-preserving transformations to examine sensitivity to identifier changes and other structural edits [23]. The robustness profiles address these complementary aspects: Figure 5(a) plots mean reciprocal ranks for five perturbations, with baseline values 0.731, 0.706, 0.688, 0.654, and 0.621 and joint values 0.814, 0.796, 0.781, 0.749, and 0.708; Figure 5(b) gives a single heat map of graph availability by language and perturbation, ranging from 0.68 to 0.98. The perturbations are clean input, identifier renaming, temporary expansion, library substitution, and partial truncation. The joint drop under truncation is 0.106, compared with 0.110 for the baseline, which corresponds to a modest reduction in degradation under this condition. Library substitution reduces the joint score by 0.065 even though token overlap may change sharply. The graph availability matrix helps expose a competing explanation: if graph extraction fails disproportionately in one language, the model may rely more often on its token branch. The robustness protocol retains relevance labels for transformations with verified semantic preservation and records failed transformations separately. Partial truncation is an incomplete-context stress test and can remove behavior-defining statements. Its score therefore measures retrieval under missing context, whereas a semantics-preserving transformation supports an invariance interpretation only after verification.

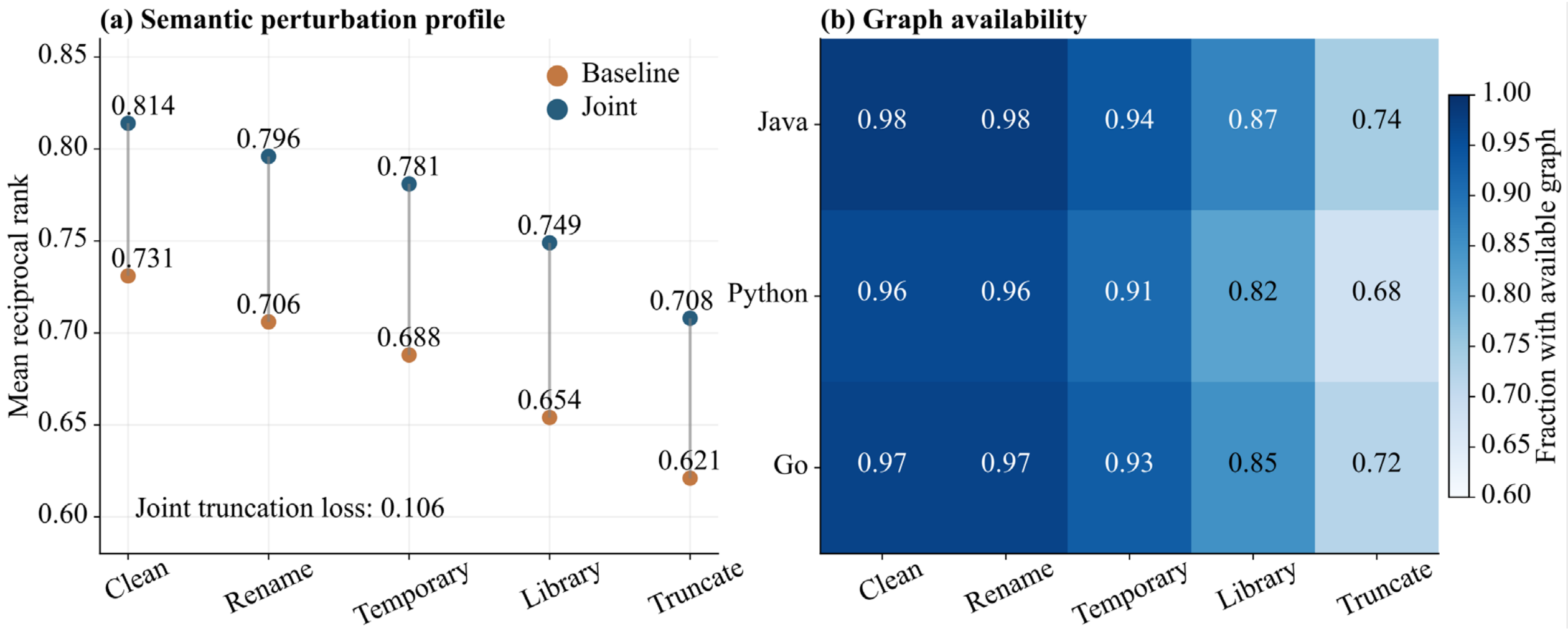

**Figure 5. Robustness analysis: (a) perturbation score scatter profile with paired connectors; (b) graph availability heat map for three languages and five perturbations**

The validation profiles characterize parameter sensitivity: Figure 6(a) shows alignment coefficients 0.05, 0.10, 0.15, 0.20, and 0.30 with scores 0.793, 0.807, 0.816, 0.810, and 0.792; Figure 6(b) shows accepted hard fractions 0.20, 0.40, 0.60, 0.80, and 1.00 with scores 0.788, 0.805, 0.816, 0.809, and 0.796. The two profiles hold the other selected parameter fixed. Evaluation of parameter-efficient adaptation for software engineering tasks examines both effectiveness and efficiency and analyzes sensitivity to a scaling-factor hyperparameter [24]. The present parameter comparison therefore assesses retrieval quality directly on the validation partition. The interior maxima of the two validation profiles identify an operating region that balances structural pressure and negative difficulty. These conditional profiles describe the response to each parameter with its counterpart fixed; a joint search would also characterize interactions. The 0.816 validation score differs from the 0.814 test mean and corresponds to a separate evaluation partition. Reporting all evaluated settings provides a complete view of the parameter search and clarifies the width of the favorable operating region. The two curves are conditional slices through the tested settings, not a complete two-dimensional search. Their maxima identify a useful validation setting within the reported grid, while interactions outside those slices remain unresolved.

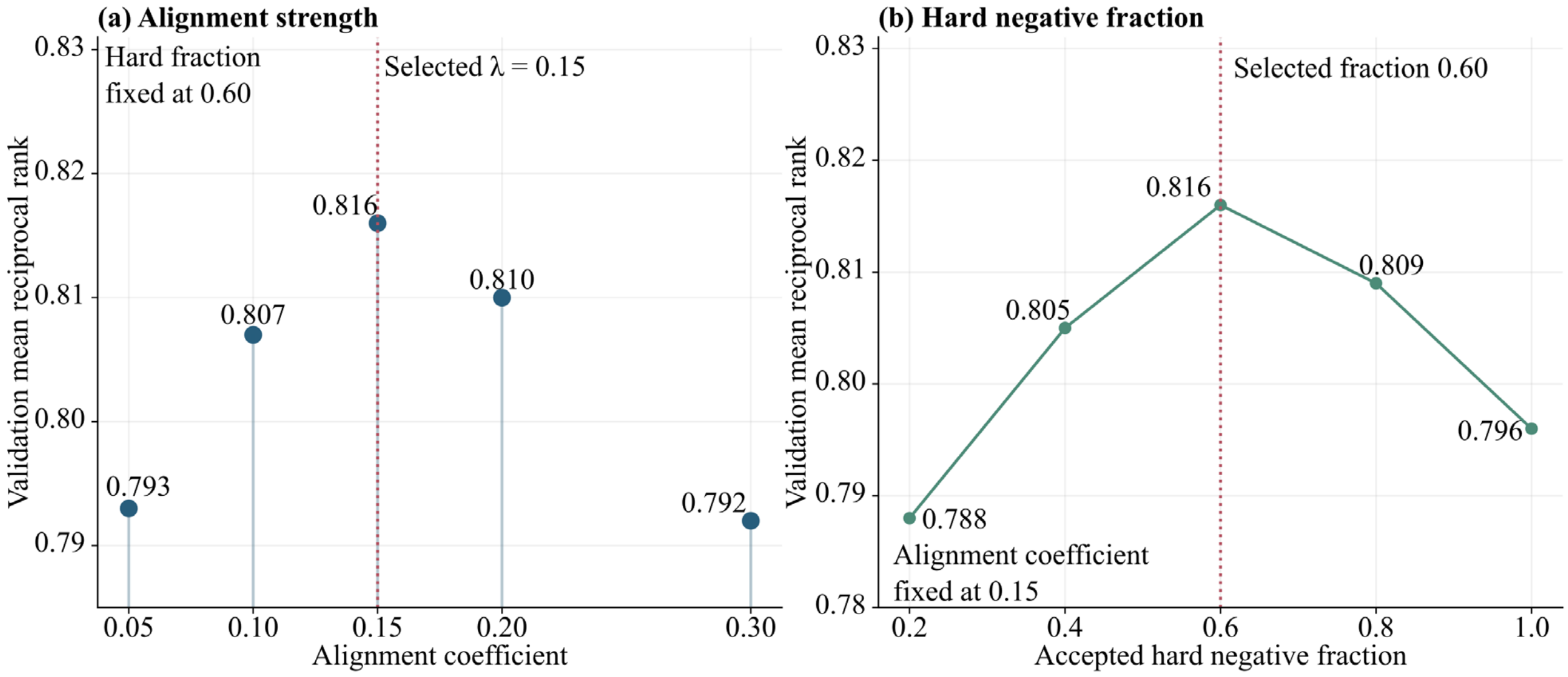

**Figure 6. Validation sensitivity: (a) alignment coefficient profile; (b) hard negative fraction profile**

Code retrieval architectures can explicitly trade inference cost against accuracy through self-distillation and intermediate encoder exits [25]. The operational analysis considers four complementary measurements: Figure 7(a) places five index settings at latency and recall pairs of (8, 0.930), (11, 0.955), (15, 0.974), (22, 0.989), and (34, 0.997), where latency is milliseconds and recall concerns agreement with exact vector search; Figure 7(b) stacks parser, encoder, lookup, and verification times for five workloads; Figure 7(c) traces accepted hard candidate yield from 0.31 to 0.68 over five refresh stages while quarantine falls from 0.22 to 0.10; Figure 7(d) relates five vector dimensions to memory footprints of 0.30, 0.43, 0.56, 0.82, and 1.07 gigabytes. The operational profile concerns a million-candidate serving configuration, evaluated separately from the smaller effectiveness collection. The memory accounting includes vector storage, identifiers, and index overhead. The selected index setting has 15 millisecond lookup latency, but a moderate workload adds 4 milliseconds of parsing, 18 of encoding, and 7 of verification, producing 44 milliseconds end to end. Omitting those stages would understate latency by nearly a factor of three. Cross platform cost comparisons require device specifications, concurrency settings, and consistent timing boundaries. Index agreement concerns preservation of exact-search neighbors, not task relevance. The 44 millisecond total applies to the stated stage sum; without device, precision, concurrency, and percentile timing records it is not a transferable service-level guarantee. For one million float32 vectors, raw storage is 1.024 decimal gigabytes at 256 dimensions and 0.512 at 128 dimensions. The corresponding profile totals of 1.07 and 0.56 gigabytes are compatible with additional overhead; reproduction still requires the actual precision, index encoding, and accounting boundary.

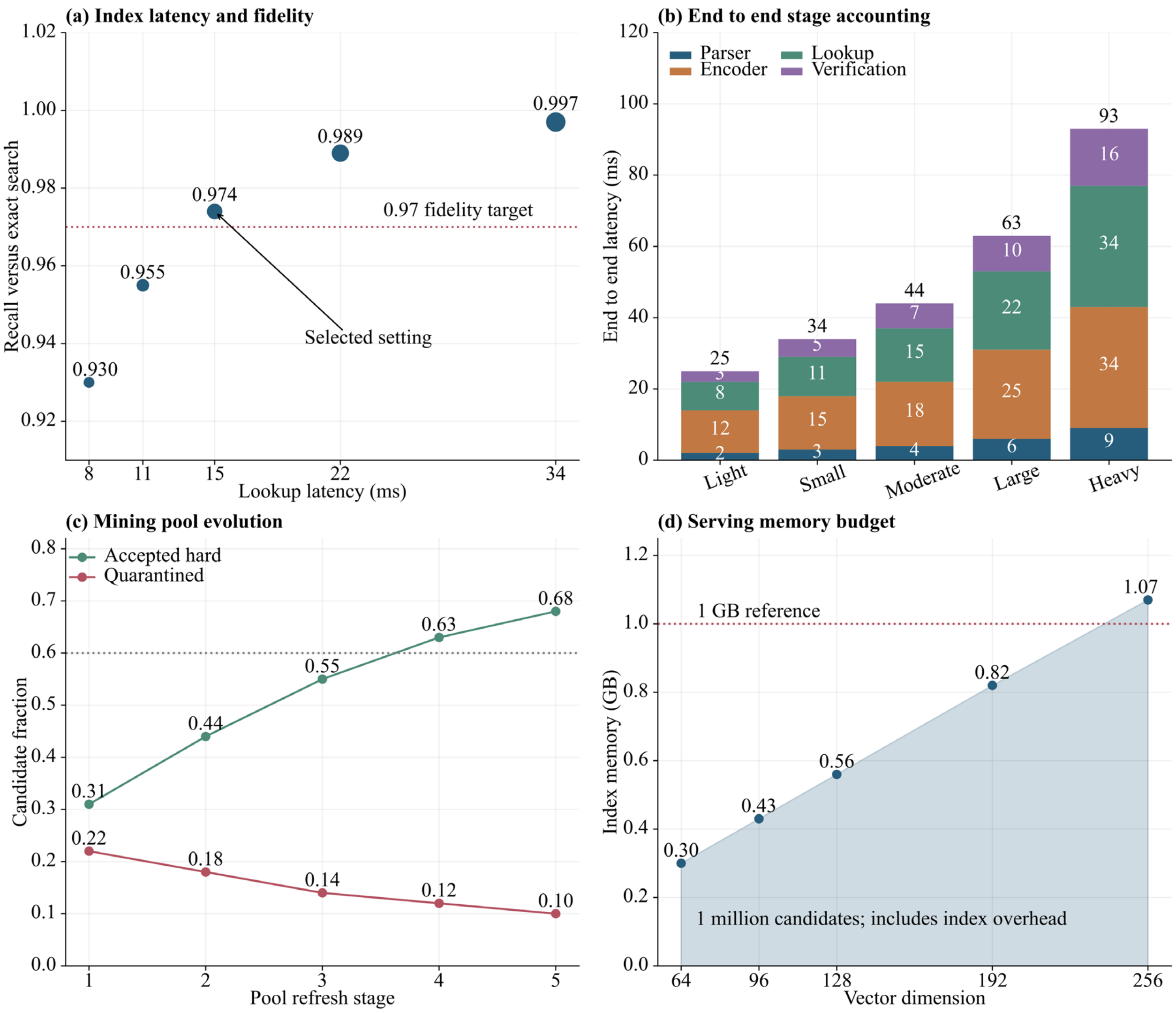

**Figure 7. Operational profiles: (a) latency versus exact search agreement scatter; (b) stacked stage latency bars; (c) mining yield curves; (d) memory footprint area profile**

Recent clone detection research combines behavior-preserving code revisions with abstract syntax tree graph modeling and contrastive supervision to address syntactically diverse semantic clones [26]. Figure 8 presents a horizontal lollipop plot with six mutually exclusive primary error counts: boundary conditions 24, library semantics 21, aliasing 18, incomplete graphs 16, task label ambiguity 12, and truncation 9. Their total is 100, making the counts directly interpretable as percentages within the error sample. The audit protocol specifies random selection from failed queries, secondary labels for overlapping causes, and independent review followed by disagreement resolution. Boundary errors should be supported by divergent behavior on an edge case; library errors require a documented semantic difference; incomplete graph errors require evidence that a missing dependency affected ranking. Category attribution requires the corresponding program evidence and distinguishing test cases. The proposed audit would retain the query, highest ranked distractor, first relevant candidate, parser record, and minimal distinguishing test for every reviewed case, allowing later investigators to challenge a category assignment and inspect failures that aggregate ranking scores conceal. These percentages are conditional on the 100 reviewed failed queries, not on all test queries. A primary label makes the six counts mutually exclusive for reporting, while secondary labels retain overlapping explanations and should be recorded separately. The 20-case marker is a descriptive prioritization aid for error review; it is not a statistical threshold or a criterion for accepting a category assignment.

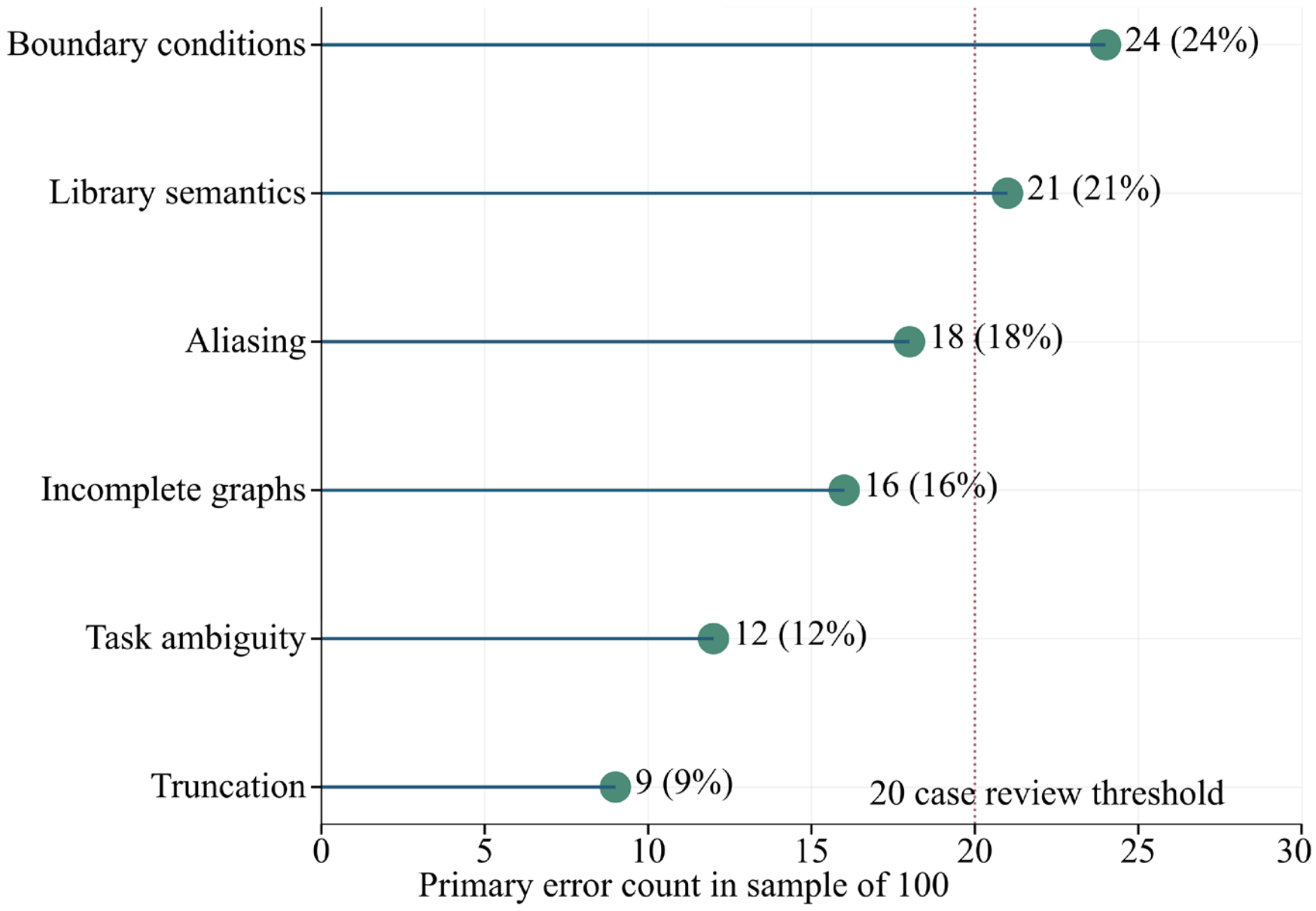

**Figure 8. Primary error taxonomy with six category counts, count annotations, and a 20-case review threshold**

## 6. Conclusion

This study develops a GraphCodeBERT model that combines soft data flow alignment with reliability filtered hard negative mining, with the evaluation indicating complementary contributions to cross language code clone retrieval.

1. Overall retrieval performance. Across six directed retrieval settings involving Java, Python, and Go, the joint configuration achieves a mean reciprocal rank of 0.814, compared with 0.731 for the matched baseline, representing an absolute increase of 0.083. Recall at ten increases from 0.682 to 0.758, while recall at fifty rises from 0.853 to 0.908. These differences indicate improvements in both ranking quality and relevant candidate coverage.

2. Component contributions and operating conditions. Alignment alone and filtered mining alone produce mean reciprocal ranks of 0.772 and 0.786, respectively, while their combination reaches 0.814. Replacing reliability filtered mining with unfiltered mining reduces the joint score to 0.794, highlighting the importance of candidate reliability alongside difficulty. Validation performance peaks at an alignment coefficient of 0.15 and a hard negative fraction of 0.60, indicating that moderate structural supervision and selective difficulty provide a favorable balance. These settings are the best among the reported conditional validation profiles; they are not established optima for other datasets, candidate distributions, or parser implementations.

3. Methodological contribution and limitations. The model integrates flexible dependency correspondence with controlled negative selection while preserving independent candidate encoding for indexed retrieval. Its effectiveness remains sensitive to structural extraction quality and semantic ambiguity. Under partial truncation, the joint mean reciprocal rank decreases from 0.814 to 0.708, demonstrating the continuing influence of incomplete program context. Library semantics, aliasing, and uncertain task labels also require additional verification.

The proposed framework provides a basis for combining structural supervision with discriminative sampling; further development should strengthen semantic verification and evaluate transfer across broader repository domains.

## 7. Declarations

### 7.1 Author Contributions

Qiong Liu contributes to conceptualization, methodology, software, validation, analysis, investigation, data collection, draft preparation, manuscript editing, visualization, supervision, project administration, and funding acquisition. All authors have read and agreed with the manuscript before its submission and publication.

### 7.2 Funding and Acknowledgments

This research received no specific financial support from any funding agency.

### 7.3 Institutional Review Board Statement

Not applicable.

### 7.4 Informed Consent Statement

Not applicable.

### 7.5 Date Availability

Not applicable.

### 7.6 Declaration of Competing Interest

The authors declare that they have no known competing financial interests or personal relationships that could have appeared to influence the work reported in this paper.

### 7.7 Declaration of generative AI and AI-assisted technologies

During the preparation of this manuscript, the authors utilized Grammarly for linguistic polishing, logical organization and draft revision. The generative AI tool was not involved in research design, data collection, data analysis, or the equation of core conclusions. All outputs generated by AI were reviewed, revised and verified manually by the authors. The authors take full responsibility for all content of this manuscript.

## 8. References

[1] Swilam, Z., Hamdy, A., & Pester, A. (2025). Improving code semantics learning using enhanced Abstract Syntax Tree. International Journal of Computers and Applications, 47(1), 57-69. https://doi.org/10.1080/1206212x.2024.2443506

[2] Zhang, Z., & Saber, T. (2025). Machine learning approaches to code similarity measurement: A systematic review. IEEE Access, 13, 51729–51764. https://doi.org/10.1109/access.2025.3553392

[3] Ahmed, G. A., Patten, J. V., Han, Y., Lu, G., Hou, W., Gregg, D., Buckley, J., & Chochlov, M. (2024). Nearest-neighbor, BERT-based, scalable clone detection: A practical approach for large-scale industrial code bases. Software: Practice and Experience, 54(12), 2349–2374. https://doi.org/10.1002/spe.3355

[4] Fang, C., Sun, W., Chen, Y., Chen, X., Wei, Z., Zhang, Q., You, Y., Luo, B., Liu, Y., & Chen, Z. (2024). ESALE: Enhancing code-summary alignment learning for source code summarization. IEEE Transactions on Software Engineering, 50(8), 2077–2095. https://doi.org/10.1109/tse.2024.3422274

[5] Feng, R., Pearce, H., Liguori, P., & Sui, Y. (2025). CGP-Tuning: Structure-aware soft prompt tuning for code vulnerability detection. IEEE Transactions on Software Engineering, 51(9), 2533–2548. https://doi.org/10.1109/tse.2025.3591934

[6] Wu, Y., Feng, S., Suo, W., Zou, D., & Jin, H. (2024). Goner: Building tree-based N-gram-like model for semantic code clone detection. IEEE Transactions on Reliability, 73(2), 1310–1324. https://doi.org/10.1109/tr.2023.3312294

[7] Jiang, S., Fu, C., He, S., Lv, J., Han, L., & Hu, H. (2024). BinCola: Diversity-sensitive contrastive learning for binary code similarity detection. IEEE Transactions on Software Engineering, 50(10), 2485–2497. https://doi.org/10.1109/tse.2024.3411072

[8] Yang, Y., Hu, X., Gao, Z., Chen, J., Ni, C., Xia, X., & Lo, D. (2024). Federated learning for software engineering: A case study of code clone detection and defect prediction. IEEE Transactions on Software Engineering, 50(2), 296–321. https://doi.org/10.1109/tse.2023.3347898

[9] Du, Y., Ma, T., Wu, L., Zhang, X., & Ji, S. (2024). AdaCCD: Adaptive semantic contrasts discovery based cross lingual adaptation for code clone detection. Proceedings of the AAAI Conference on Artificial Intelligence, 38(16), 17942–17950. https://doi.org/10.1609/aaai.v38i16.29749

[10] Sorokin, N., Anton, T., Abulkhanov, D., Sedykh, I., Piontkovskaya, I., & Malykh, V. (2025). CCT-Code: Cross-consistency training for multilingual clone detection and code search. In A. Ebrahimi, S. Haider, E. Liu, S. Haider, M. L. Pacheco, & S. Wein (Eds.), Proceedings of the 2025 Conference of the Nations of the Americas Chapter of the Association for Computational Linguistics: Human Language Technologies (Vol. 4: Student Research Workshop, pp. 178–185). Association for Computational Linguistics. https://doi.org/10.18653/v1/2025.naacl-srw.17

[11] Wu, Y., Suo, W., Feng, S., Wu, C., Zou, D., & Jin, H. (2025). Fine-grained code clone detection by keywords-based connection of program dependency graph. IEEE Transactions on Reliability, 74(3), 3427–3441. https://doi.org/10.1109/tr.2025.3550747

[12] Hong, J., Choi, E., & Mizuno, O. (2024). GNEM: Comprehensive similarity learning with ensemble model for code search. IEEE Access, 12, 165005–165016. https://doi.org/10.1109/access.2024.3493193

[13] Fan, Y., Li, C., Ge, J., Huang, L., & Luo, B. (2025). Effective hard negative mining for contrastive learning-based code search. ACM Transactions on Software Engineering and Methodology, 34(3), 1–35. https://doi.org/10.1145/3695994

[14] Liu, H., Zhan, J., & Zhang, Q. (2025). Uncertainty-aware contrastive learning with hard negative sampling for code search tasks.

Proceedings of the AAAI Conference on Artificial Intelligence, 39(18), 18807–18815. https://doi.org/10.1609/aaai.v39i18.34070

[15] Hernández López, J. A., Chen, B., Saad, M., Sharma, T., & Varró, D. (2025). On inter-dataset code duplication and data leakage in large language models. IEEE Transactions on Software Engineering, 51(1), 192–205. https://doi.org/10.1109/tse.2024.3504286

[16] Gong, J., Wu, Y., Liang, L., Wang, Y., Chen, J., Liu, M., & Zheng, Z. (2026). CoSQA+: Enhancing code search evaluation with a multi-choice benchmark and test-driven agents. IEEE Transactions on Software Engineering, 52(1), 206–220. https://doi.org/10.1109/tse.2025.3631886

[17] Li, X., Dong, K., Lee, Y. Q., Xia, W., Zhang, H., Dai, X., Wang, Y., & Tang, R. (2025). CoIR: A comprehensive benchmark for code information retrieval models. In W. Che, J. Nabende, E. Shutova, & M. T. Pilehvar (Eds.), Proceedings of the 63rd Annual Meeting of the Association for Computational Linguistics (Vol. 1: Long Papers, pp. 22074–22091). Association for Computational Linguistics. https://doi.org/10.18653/v1/2025.acl-long.1072

[18] Zou, W., Li, Q., Li, C., Ge, J., Chen, X., Huang, L., & Luo, B. (2025). Improving source code pre-training via type-specific masking. ACM Transactions on Software Engineering and Methodology, 34(3), 1–34. https://doi.org/10.1145/3699599

[19] Moumoula, M. B., Kaboré, A. K., Klein, J., & Bissyandé, T. F. (2025). The struggles of LLMs in cross-lingual code clone detection. Proceedings of the ACM on Software Engineering, 2(FSE), 1023–1045. https://doi.org/10.1145/3715764

[20] Fan, G., Chen, S., Gao, C., Xiao, J., Zhang, T., & Feng, Z. (2024). RAPID: Zero-shot domain adaptation for code search with pre-trained models. ACM Transactions on Software Engineering and Methodology, 33(5), 1–35. https://doi.org/10.1145/3641542

[21] Glani, Y., Ping, L., Shah, S. A., & Ke, L. (2025). CCDive: A deep dive into code clone detection using local sequence alignment. Tsinghua Science and Technology, 30(4), 1435–1456. https://doi.org/10.26599/tst.2024.9010075

[22] Feng, J., Li, W., Wu, S., Wei, Z., Xu, Y., Wang, J., & Li, H. (2025). Deep code search with naming-agnostic contrastive multi-view learning. ACM Transactions on Knowledge Discovery from Data, 19(6), 1–26. https://doi.org/10.1145/3737878

[23] Li, Y., Qi, S., Gao, C., Peng, Y., Lo, D., Lyu, M. R., & Xu, Z. (2025). Understanding the robustness of Transformer-based code intelligence via code transformation: Challenges and opportunities. IEEE Transactions on Software Engineering, 51(2), 521–547. https://doi.org/10.1109/tse.2024.3524461

[24] Zou, W., Shen, Z., Li, Q., Ge, J., Li, C., Chen, X., Shen, X., Huang, L., & Luo, B. (2025). Experimental evaluation of parameter-efficient fine-tuning for software engineering tasks. ACM Transactions on Software Engineering and Methodology, 34(7), 1–34. https://doi.org/10.1145/3722107

[25] Gurioli, A., Pennino, F., Monteiro, J., & Gabbrielli, M. (2026). MoSE: Hierarchical self-distillation enhances early layer embeddings. Proceedings of the AAAI Conference on Artificial Intelligence, 40(37), 30897–30906. https://doi.org/10.1609/aaai.v40i37.40348

[26] Li, C., Konpang, J., Sirikham, A., & Wang, Y. (2025). Nuanced code clone detection through LLM-based code revision and AST graph modeling. IEEE Access, 13, 191024–191036. https://doi.org/10.1109/access.2025.3628856